%% file: main.tex
\PassOptionsToPackage{table}{xcolor}
\documentclass[11pt]{article}

\usepackage[preprint]{acl}
\usepackage{times}
\usepackage{latexsym}
\usepackage[T1]{fontenc}
\usepackage[utf8]{inputenc}
\usepackage{microtype}
\usepackage{inconsolata}
\usepackage{amsmath,amssymb,amsthm,booktabs,graphicx}
\usepackage{algorithm}
\usepackage{enumitem}
\usepackage{tcolorbox}
\usepackage[noend]{algpseudocode}
\usepackage{multirow}
\definecolor{cordrow}{RGB}{238,236,250}

\newtheorem{proposition}{Proposition}
\newtheorem{lemma}[proposition]{Lemma}

\newcommand{\E}{\mathbb{E}}
\newcommand{\TV}{\mathrm{TV}}
\newcommand{\METHOD}{\textsc{CoRD}}

\title{CoRD: Coupled Rollouts Fill the Blind Spot of On-Policy Distillation}
\author{
 \textbf{Guhan Chen\textsuperscript{1}\thanks{Equal Contribution.},~}
 \textbf{Bohan Li\textsuperscript{1$\ast$},~}
 \textbf{Jun Rao\textsuperscript{2},~} 
 \textbf{Hejin Wang\textsuperscript{1},~}
 \textbf{Songtao Tian\textsuperscript{1},~} 
 \textbf{Yexin Xie\textsuperscript{3},~}\\
 \textbf{Xiaojun Meng\textsuperscript{4},~}
 \textbf{Peng Yuan\textsuperscript{4},~}
 \textbf{Jiansheng Wei\textsuperscript{4},~}
  \textbf{Xuebo Liu\textsuperscript{2},~}
 \textbf{Zixiong Yu\textsuperscript{4}\thanks{Corresponding Author.}}
\\
 \textsuperscript{1}Tsinghua University\quad
 \textsuperscript{2}{Harbin Institute of Technology, Shenzhen, China}\\
 \textsuperscript{3}Nanjing University\quad
 \textsuperscript{4}Huawei Large Model Data Technology Lab\\
\texttt{chen-gh23@mails.tsinghua.edu.cn}\quad\texttt{rao7jun@gmail.com}\\
\texttt{\{libh19,wanghj20,tiansongtao.2020,yuzx19\}@tsinghua.org.cn}\\
\texttt{\{xiaojun.meng,yuanpeng126,weijiansheng\}@huawei.com}\\
\texttt{xieye-xin@qq.com}\quad
\texttt{liuxuebo@hit.edu.cn}%
}

\begin{document}
\maketitle
\begin{abstract}
\input{sections/abstract}

\end{abstract}

\input{sections/intro}
\input{sections/related}
\input{sections/method}
\input{sections/experiments}
\input{sections/analysis}
\input{sections/conclusion}
\input{sections/limitations}

\bibliography{references}

\appendix
\input{appendix/details}
\input{appendix/sampler}

\end{document}

%% file: sections/abstract.tex
On-policy distillation trains a student on its own rollouts, so states the teacher would reach but the student does not are seldom supervised. We introduce \METHOD{} (coupled-rollout distillation), which keeps the on-policy loss and moves only the distribution of training states toward the teacher, drawing rollouts from a geometric interpolation of student and teacher through the maximal coupling of speculative decoding. The coupling keeps the student's token wherever the two distributions agree and otherwise substitutes a token from the part of the target the student underweights, so interventions are as rare as any exact sampler allows and recede as the student learns. Across six base students from two model families (Qwen3 and Qwen3.5, 0.6B to 4B), \METHOD{} raises average pass@1 over on-policy distillation by 0.9 to 2.0 points and pass@8 by 1.2 to 3.4 points on five public math benchmarks at equal steps and data. On frozen teacher-reached states, a 4B student trained with \METHOD{} reaches in 16 steps a gap closure that on-policy distillation does not reach in 80, and substitution rules intervening at least as often in other directions fall short of \METHOD{}.

%% file: sections/intro.tex
\section{Introduction}

The rapid advancement of artificial intelligence has been driven by complementary progress in model architectures \citep{he2016deep,vaswani2017attention,yu2026branch}, data scale and quality \citep{yuan2023scalingrelationshiplearningmathematical,rao2025data,yu-etal-2026-mathagent}, and training paradigms \citep{ouyang2022training}. Current research frontiers span theoretical explorations of learning dynamics and generalization \citep{chen2024OntheImpacts,li2024eigenvalue,yu2025divergence}, generation across various modalities \citep{yu2024samwav2lip,tian2026curvatureadaptiveconsistencyflowmatching}, tool-augmented reasoning \citep{wu2026toolaugmentedpolicyoptimizationsynergizing}, and interdisciplinary applications \citep{liu2026llmstacklehighenergy}. Within this broad landscape, large language models (LLMs) have emerged as a central paradigm for exploring complex reasoning and knowledge transfer. A key challenge is how to enable smaller student models to efficiently inherit the capabilities of stronger teachers.

\begin{figure*}[t]
\centering
\includegraphics[width=0.9\textwidth]{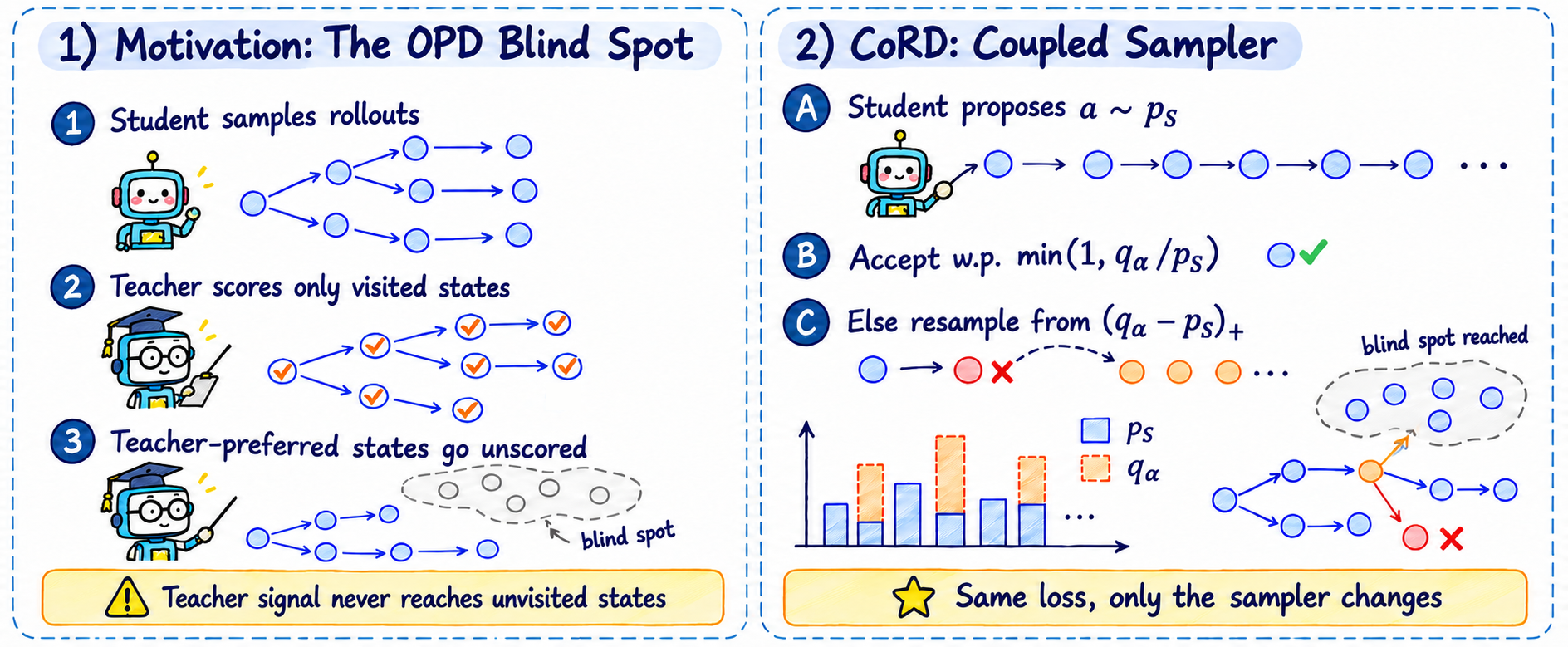}
\caption{\textbf{Overview of \METHOD{}.} (i)~\textit{Left (OPD's blind spot):} On-policy distillation trains only on student rollouts. Teacher-preferred reasoning paths that the student rarely or never visits therefore receive little or no supervision. (ii)~\textit{Right (our coupled sampler):} \METHOD{} keeps the same training loss and budget, but changes the rollout sampler. By coupling the student with a teacher-leaning target $q_\alpha$, it preserves agreeing tokens and replaces disagreements with underweighted target tokens. This maximal coupling steers rollouts into teacher-preferred states missed by OPD with the minimum possible intervention rate, minimally perturbing the student's original trajectory.}
\label{fig:teaser}
\end{figure*}

Among knowledge distillation methods, on-policy distillation (OPD) has become a commonly used approach to this challenge \cite{agarwal2024gkd, gu2024minillm, yang2025qwen3, xu2024kdsurvey}. The student generates its own rollouts and is trained to match the teacher's full next-token distribution at every state it visits, usually by minimizing the reverse Kullback--Leibler divergence. Training on self-generated text removes the mismatch between teacher-forced training and free-running generation \cite{bengio2015scheduled, ranzato2016sequence, ross2011dagger}, and the teacher's dense, token-level signal replaces the single final reward of reinforcement learning.

The same design also fixes where the supervision lands. Because states are drawn from the student, the teacher only ever comments on states the student already reaches. States the teacher would enter, such as the first step of a solution path the student considers unlikely, are sampled only as often as the student itself picks them, and corrected only that often. We measure this blind spot on frozen sets of states. Before training, the student is 1.5 to 2.4 times further from the teacher on states reached by the teacher's distribution than on states it reaches itself, across the four students analyzed in Section~\ref{sec:analysis}. On-policy training closes this gap only slowly: after 40 steps at 4B, OPD has removed 15.0\% of the divergence on teacher-reached states, where the sampler we introduce removes 21.7\%.

Reaching those states requires steering the rollout toward the teacher while leaving it the student's own text wherever the two agree. Speculative decoding already contains the right tool \cite{leviathan2023speculative, chen2023speculative}. Its rejection sampler, usually seen as an accelerator, realizes the maximal coupling of two token distributions \cite{lindvall1992coupling}: a draft token is kept where the distributions agree and replaced by a draw from their difference where they do not. When used instead as a training sampler, this rule intervenes only where the student and the target disagree, and exactly as often as they disagree.

We introduce \METHOD{} (coupled-rollout distillation), illustrated in Figure~\ref{fig:teaser}, which changes the rollout sampler of OPD and nothing else. At each position the student proposes a token, and the coupling with a tempered target $q_\alpha \propto p_{\mathrm{S}}^{1-\alpha} p_{\mathrm{T}}^{\alpha}$ either keeps it or substitutes a token from the residual mass that the student underweights. Every emitted token is an exact draw from $q_\alpha$, the probability of an intervention equals the total-variation distance between $p_{\mathrm{S}}$ and $q_\alpha$, and $\alpha=0$ recovers OPD bit for bit. The loss, the teacher, and the number of training steps stay as they are; only the distribution of training states moves toward the teacher. That shift is the source of the gain. The coupling is how \METHOD{} reaches the shifted states while rewriting as little of the student's own text as any exact sampler of $q_\alpha$ can, and it splits every draw into an acceptance decision and a replacement direction, which Section~\ref{sec:analysis} tests separately.

The shift pays off on public benchmarks. Across six base students from two model families, \METHOD{} improves average pass@1 and pass@8 over OPD on MATH500, AIME24, AIME25, AMC23 and GSM8K for every student in Table~\ref{tab:public}; on Qwen3.5-0.8B and Qwen3.5-2B, SKD's published replacement rule \cite{xu2025skd}, with the same loss and budget, lands between OPD and \METHOD{}, and OPD trained for twice as many steps does not catch up. The difficulty breakdown shows that each student's largest pass@16 gain is on problems it rarely or never solved before training, although the aggregate pass@16 gain is not positive for every student. Substitution rules that intervene at least as often but draw the replacement from elsewhere, namely a uniform token, a token from the student's average distribution, or a teacher sample at randomly chosen positions, stay at or near OPD and far below \METHOD{}, which ties the gain to where the rollouts are steered rather than to the act of intervening.

Our contributions are as follows:
\begin{itemize}[leftmargin=*,nosep]
\item We measure the blind spot of on-policy distillation on frozen sets of teacher-reached states, where the student starts 1.5 to 2.4 times further from the teacher than on its own states, and show that OPD leaves most of it in place.
\item We propose \METHOD{}, which trains on rollouts from a teacher-leaning target and realizes them with the speculative-decoding rejection sampler: exact draws from the target with the fewest possible replacements (the total-variation distance), and OPD as the special case $\alpha=0$.
\item We show gains in average pass@1 and pass@8 over OPD for all six base students on five public math benchmarks, and study teacher-reached states through closure measurements, substitution controls and an $\alpha$ sweep.
\end{itemize}

%% file: sections/related.tex
\section{Related Work}

\paragraph{Knowledge Distillation for Language Models.}
Distillation trains a student to match a teacher's output distribution \cite{hinton2015distilling, rao2024pesfkd, jin2026student}, extended to sequences generated by the teacher \cite{kim2016sequence} and to other divergences \cite{wen2023fdivergence}; \citet{xu2024kdsurvey} survey the area. On-policy methods evaluate the loss on the student's own samples \cite{lin2020autoregressive, agarwal2024gkd, fu2026revisitingopd, rao2026transfer} and now train production models \cite{yang2025qwen3, lu2025onpolicy}. DistiLLM \cite{ko2024distillm, ko2025distillm2} mixes fixed and student-generated outputs under a skewed divergence, and TAID \cite{shing2025taid} moves the training target from the student toward the teacher over training. These methods change the loss, its target, or the data mix \cite{chen2026diag, rao2026sai, rao2026pace}; \METHOD{} keeps the on-policy loss and target and changes only where the states come from, so it applies unchanged to any per-state loss computed from both distributions.

\paragraph{Exposure Bias and Imitation Learning.}
Training on states the model would not visit at test time is the exposure-bias problem \cite{bengio2015scheduled, ranzato2016sequence}. Scheduled sampling mixes reference and model tokens under a hand-set schedule \cite{bengio2015scheduled}; SEARN, DAgger and MiniLLM roll out a mixture of expert and learner \cite{daume2009searn, ross2011dagger, gu2024minillm}, the last with importance weights. Their mixing follows a schedule or a fixed probability; in \METHOD{} the probability of intervening at a token is the current disagreement itself, so no schedule is needed and interventions vanish as student and target agree. Section~\ref{sec:analysis} includes a DAgger-style control with the same average rate.

\paragraph{Speculative Decoding and Interleaved Sampling.}
Speculative decoding lets a draft model propose tokens that a rejection sampler accepts or replaces so that the output follows the target exactly \cite{leviathan2023speculative, chen2023speculative}; the sampler is a maximal coupling \cite{lindvall1992coupling}. DistillSpec \cite{zhou2024distillspec} uses distillation to improve the draft; \METHOD{} conversely uses the sampler to improve distillation. Other recent work lets the teacher take over parts of the rollout when the student drifts \cite{xu2026relayopd, wang2026motab, liu2026tgpo}, draws rollouts from a weaker negative policy \cite{hwang2026npopd}, or repairs the OPD objective itself \cite{fu2026revisitingopd}; none samples from an explicit target with an exact rejection rule. The closest prior work is SKD \cite{xu2025skd}, where a student proposal outside the teacher's top-$K$ tokens is replaced by a teacher sample. Its replacement can be any teacher token, including ones the student already overweights, and its acceptance set is fixed by $K$, so its intervention rate is not tied to the distance between the two distributions. \METHOD{} replaces only from the residual $(q_\alpha - p_{\mathrm{S}})_+$, emits tokens that follow $q_\alpha$ exactly, intervenes at the rate $\TV(p_{\mathrm{S}}, q_\alpha)$, and moves continuously from OPD ($\alpha=0$) to teacher-weighted states ($\alpha=1$). Section~\ref{sec:skd} compares the two under the same loss.

%% file: sections/method.tex
\section{Method}
\label{sec:method}

\paragraph{On-Policy Distillation.}
A student policy $p_{\mathrm{S}}(\cdot \mid s)$ is trained toward a frozen teacher $p_{\mathrm{T}}(\cdot \mid s)$, where a state $s$ is a prompt followed by a partial response. OPD minimizes the reverse KL divergence at the states visited by the student's own rollouts,
\begin{equation}
\mathcal{L}_{\text{OPD}} = \E_{s \sim d_{p_{\mathrm{S}}}} \big[ D_{\mathrm{KL}}\big(p_{\mathrm{S}}(\cdot \mid s) \,\|\, p_{\mathrm{T}}(\cdot \mid s)\big) \big],
\label{eq:opd}
\end{equation}
where $d_{p_{\mathrm{S}}}$ is the distribution of states induced by sampling from $p_{\mathrm{S}}$ and the divergence is computed over the full vocabulary $\mathcal{V}$. The loss supervises every visited state densely, but the visited states come from $d_{p_{\mathrm{S}}}$ alone.

\paragraph{A Teacher-Leaning Target.}
\METHOD{} samples rollouts from the tempered target
\begin{equation}
q_\alpha(v \mid s) = \frac{p_{\mathrm{S}}(v \mid s)^{1-\alpha}\, p_{\mathrm{T}}(v \mid s)^{\alpha}}{\sum_{u \in \mathcal{V}} p_{\mathrm{S}}(u \mid s)^{1-\alpha}\, p_{\mathrm{T}}(u \mid s)^{\alpha}},
\label{eq:target}
\end{equation}
a geometric interpolation that equals the student at $\alpha=0$ and the teacher at $\alpha=1$. Computing $q_\alpha$ needs both full logit vectors at the current position, which the loss in Eq.~\eqref{eq:opd} uses anyway.

\begin{algorithm}[t]
\caption{\METHOD{} rollout for one prompt}
\label{alg:cord}
\small
\begin{algorithmic}[1]
\Require student $p_{\mathrm{S}}$, teacher $p_{\mathrm{T}}$, prompt $x$, $\alpha \in [0,1]$
\State $s \gets x$
\While{not finished}
  \State compute $p_{\mathrm{S}}(\cdot \mid s)$, $p_{\mathrm{T}}(\cdot \mid s)$ and $q_\alpha(\cdot \mid s)$ \Comment{Eq.~\ref{eq:target}}
  \State propose $a \sim p_{\mathrm{S}}(\cdot \mid s)$; draw $u \sim \mathcal{U}[0,1]$
  \If{$u \le q_\alpha(a \mid s)/p_{\mathrm{S}}(a \mid s)$}
    \State $y \gets a$ \Comment{keep the student's token}
  \Else
    \State $y \sim (q_\alpha - p_{\mathrm{S}})_+ / \TV(p_{\mathrm{S}}, q_\alpha)$ \Comment{residual}
  \EndIf
  \State $s \gets s \oplus y$
\EndWhile
\State \Return visited states; the loss is the per-state term of Eq.~\eqref{eq:opd} at each
\end{algorithmic}
\end{algorithm}

\paragraph{The Coupled Sampler.}
Algorithm~\ref{alg:cord} gives the rollout. The student proposes $a \sim p_{\mathrm{S}}$, which is kept with probability $\min(1, q_\alpha(a)/p_{\mathrm{S}}(a))$. Otherwise a replacement is drawn from the normalized residual $(q_\alpha - p_{\mathrm{S}})_+$, the tokens that the target weights more heavily than the student does, and the rollout continues from the replacement. This is the rejection rule of speculative decoding \cite{leviathan2023speculative}, with the student as the draft and $q_\alpha$ as the target. The rollout follows $q_\alpha$ exactly: a draw from $q_\alpha$ gives the same distribution of training states and the same expected update, and this shift of the states drives the gains. The coupling determines how the shift is realized. Among all ways to pair the student's proposal with a draw from $q_\alpha$, it is the one that departs from the proposal least often, it makes every intervention explicit and measurable, and at $\alpha=0$ it reproduces OPD bit for bit.

\begin{proposition}
\label{prop:exact}
Each emitted token is an exact sample from $q_\alpha(\cdot \mid s)$, and the probability of a replacement at state $s$ equals $\TV\big(p_{\mathrm{S}}(\cdot \mid s), q_\alpha(\cdot \mid s)\big)$.
\end{proposition}

\noindent The proof is the standard one for the rejection sampler of speculative decoding: a token $v$ is emitted with probability $\min(p_{\mathrm{S}}(v), q_\alpha(v)) + (q_\alpha(v) - p_{\mathrm{S}}(v))_+ = q_\alpha(v)$, and the total rejection mass is $\sum_v (p_{\mathrm{S}}(v) - q_\alpha(v))_+ = \TV(p_{\mathrm{S}}, q_\alpha)$, the smallest probability with which any coupling of the two distributions can differ. Three consequences matter for distillation. The sampler intervenes only where the student and the target disagree, and in proportion to the disagreement, so the intervention rate is a measured quantity rather than a schedule and goes to zero as $p_{\mathrm{S}}$ approaches $p_{\mathrm{T}}$. At $\alpha=0$ the target equals the student, no proposal is ever rejected, and the rollout, the loss and the gradient are identical to OPD. And a replacement is sampled in proportion to the excess $q_\alpha(v) - p_{\mathrm{S}}(v)$. For $\alpha>0$, since $q_\alpha(v)/p_{\mathrm{S}}(v) = (p_{\mathrm{T}}(v)/p_{\mathrm{S}}(v))^{\alpha}/Z_\alpha$ with normalizer $Z_\alpha \le 1$, the excess is positive exactly on the tokens whose teacher-to-student ratio exceeds $Z_\alpha^{1/\alpha}$, a set that contains every token the teacher weights more than the student, so each intervention moves the rollout toward the tokens the teacher favors most relative to the student.

\begin{figure}[t]
\centering
\includegraphics[width=0.99\columnwidth]{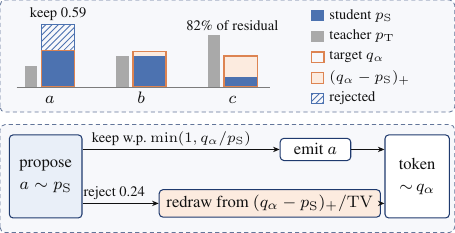}
\caption{\textbf{One-step coupled sampling.} On a three-token vocabulary with $\alpha=0.6$, student proposals are kept whenever possible and replaced only when necessary, so the emitted token follows $q_\alpha$ exactly.}
\label{fig:example}
\end{figure}

\paragraph{A Worked Example.}
In Figure~\ref{fig:example} the student overweights $a$ and the teacher prefers $c$, so the sampler rejects a proposal of $a$ with probability 0.41 and never rejects $b$ or $c$. The replacement comes from the residual, which puts 82\% of its mass on $c$ and none on $a$: the rollout is pushed toward the token the student underweights, and only as often as needed, 24\% of the time, to make the emitted token follow $q_\alpha$. A rule that replaces every token outside the teacher's top-1 with a teacher sample, as SKD does with $K=1$, would intervene 90\% of the time and could still emit $a$.

\paragraph{Objective.}
The loss remains the full-vocabulary reverse KL defined in Eq.~\eqref{eq:opd}, evaluated at states $s \sim d_{q_\alpha}$. Because the divergence is zero only when the two distributions coincide, the minimizer at every visited state is still $p_{\mathrm{S}} = p_{\mathrm{T}}$; \METHOD{} changes which states are weighted, not the per-state minimizer. During training the realized replacement rate tracks the mean total-variation distance computed from the same logits. As verified in Appendix~\ref{app:exact}, the $\alpha=0$ sampler reproduces the OPD gradient exactly.

\paragraph{No Importance Correction.}
Training on states from $d_{q_\alpha}$ rather than $d_{p_{\mathrm{S}}}$ is the point of the method, so we do not reweight the visited states back to the student's distribution. Such a correction, a product of ratios $p_{\mathrm{S}}/q_\alpha$ along the prefix, would undo the shift and its variance would grow with the length of the rollout. As in OPD, the unchanged per-state loss is applied to the visited states, the sampling distribution is treated as fixed, and the gradient flows only through $p_{\mathrm{S}}$.

\paragraph{Cost.}
The coupled rollout needs the teacher's logits at every decoding position, while OPD can generate with the student alone and score the finished rollout with one teacher forward pass. Our implementation steps both models at every position for every variant, OPD included, so all variants in our comparisons have the same cost per training step, with step times within 2.3\% of each other, and differ only in the sampling rule. Section~\ref{sec:cost} also gives OPD twice the training steps.

%% file: sections/experiments.tex
\input{sections/table_public_single_seed}

\section{Experiments}
\label{sec:experiments}

\paragraph{Models.}
We distill six base students from two model families: (i) Qwen3-0.6B-Base, Qwen3-1.7B-Base and Qwen3-4B-Base from Qwen3-8B-Base \cite{yang2025qwen3}, (ii) Qwen3.5-0.8B-Base, Qwen3.5-2B-Base and Qwen3.5-4B-Base from Qwen3.5-9B-Base \cite{qwen2026qwen35}. Students and teachers in a family share a tokenizer, which the coupling requires.

\paragraph{Training.}
Unless noted, every run trains for 80 steps on 256 prompts per step, drawn in one fixed order from a fixed set of about 5{,}000 problems of the MATH training set \cite{hendrycks2021math}, with at most 320 new tokens per rollout, learning rate $10^{-5}$ and the full-vocabulary reverse KL loss. Prompts use a fixed two-shot completion format. OPD and \METHOD{} use the same code, data, data order and number of steps, and differ only in $\alpha$: 0 for OPD and 0.6 for \METHOD{} throughout. Section~\ref{sec:alpha} shows how other values of $\alpha$ behave.

\paragraph{Evaluation.}
We evaluate students on MATH500 \cite{hendrycks2021math, lightman2024verify}, AIME24 and AIME25 \cite{aime2024, aime2025}, AMC23 \cite{amc2023} and GSM8K \cite{cobbe2021gsm8k}, drawing 16 samples per problem at temperature 1.0 with a 4096-token limit in the training prompt format. Answers are scored by the SimpleRL-Zoo math checker \cite{zeng2025simplerlzoo}, and pass@$k$ uses the unbiased estimator of \citet{chen2021codex}. No hyperparameter or checkpoint was selected on these sets.

\subsection{Main Results}
Table~\ref{tab:public} shows that \METHOD{} improves average pass@1 and pass@8 over OPD for all six students. Average pass@1 rises by 0.9 to 2.0 percentage points and average pass@8 by 1.2 to 3.4 points across the five benchmarks. Pass@16 is less uniform (Appendix Table~\ref{tab:app-pass16}): Qwen3-0.6B and Qwen3-4B gain 3.0 and 4.6 points on average, respectively, while Qwen3-1.7B is 0.1 points lower. Table~\ref{tab:difficulty} locates the largest within-student pass@16 gains: grouping all 1{,}919 problems by how many of 16 samples the untrained student solved, every student's largest gain is on problems solved at most three times initially, reaching 5.9 points on never-solved problems for Qwen3.5-4B and 6.2 points on rarely solved ones for Qwen3-0.6B. This difficulty breakdown need not imply a positive aggregate pass@16 difference for every student.

\input{sections/table_difficulty}
 The largest movements are on the benchmarks where the students have room to improve: \METHOD{} is ahead on MATH500 and GSM8K pass@1 for all six students, by up to 5.8 points on GSM8K for Qwen3-0.6B. Comparing with the untrained students shows what this shift in states buys. OPD sometimes stagnates or regresses: Qwen3-4B-Base is nearly unchanged from its untrained pass@1, and Qwen3.5-2B-Base loses about one point of average pass@8. \METHOD{} improves on the untrained student in both averages for all six. Under the same loss and budget, OPD can stall on its own states, while the coupled sampler keeps supplying states where the teacher can still teach.

\begin{figure*}[t]
\centering
\includegraphics[width=0.99\textwidth]{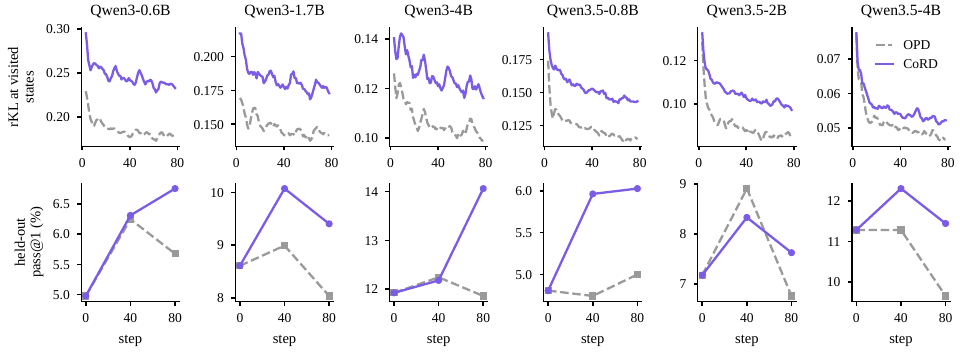}
\caption{\textbf{Visited-state divergence and held-out accuracy.} \METHOD{} maintains higher visited-state reverse KL and ends with higher held-out pass@1 for all six students in Table~\ref{tab:public}. Top: reverse KL to the teacher at states visited during training (5-step moving average). Bottom: pass@1 on about 200 held-out problems from the training distribution, evaluated at steps 0, 40 and 80.}
\label{fig:curves-public}
\end{figure*}

\subsection{Learning Curves}
Figure~\ref{fig:curves-public} shows how the two samplers train. The reverse KL at the visited states, averaged over five steps, stays higher under \METHOD{} for every student throughout the 80 steps: the coupled sampler keeps routing the rollouts through states where the student is still far from the teacher, so the loss at the visited states stays larger, whereas on OPD's own states it quickly falls. On a held-out slice of the training distribution, \METHOD{} ends above OPD for all six students.

\subsection{Comparison with SKD}
\label{sec:skd}
SKD \cite{xu2025skd} also lets the teacher intervene in the student's rollout. To compare the two sampling rules alone, we run SKD's published rule, which replaces a proposal outside the teacher's top 25 tokens by a teacher sample, with the same reverse KL loss, data, steps and per-step cost, for Qwen3.5-0.8B-Base and Qwen3.5-2B-Base; these are the rows ``SKD'' in Table~\ref{tab:public}. SKD lands between OPD and \METHOD{} on both averages for both students: it gains 0.6 and 0.2 points of average pass@1 over OPD, respectively, where \METHOD{} gains 1.1 and 0.9 points. The two rules differ in how often, where and with what they intervene. With $K=25$, SKD replaces fewer than 1\% of the tokens, 0.7\% and 0.6\% on average over training, because the student's proposal almost always falls inside the teacher's top 25 even where the two distributions differ substantially; \METHOD{} replaces 6.8\% and 5.7\%, exactly the total-variation distance to its target. A top-$K$ rule asks whether the student's token is plausible to the teacher, and where both models spread their mass over many reasonable continuations it accepts almost every proposal; the coupling moves exactly the excess of the target over the student, so supervision shifts toward the teacher wherever the two distributions differ, in proportion to how much they differ.

\subsection{OPD with Twice the Steps}
\label{sec:cost}
Because the coupled rollout queries the teacher during generation, we also give OPD twice the training steps: we train OPD on Qwen3.5-2B-Base for 160 steps, twice as many as \METHOD{}. The extra steps do not help. The corresponding ``OPD, 160 steps'' row in Table~\ref{tab:public} has average pass@1 of 15.1 and pass@8 of 35.6, below the 80-step OPD values of 16.0 and 35.7 and the 80-step \METHOD{} values of 16.9 and 37.5.

%% file: sections/table_public_single_seed.tex
\begin{table*}[t]
\centering
\setlength{\tabcolsep}{2.5pt}
\resizebox{0.99\textwidth}{!}{%
\begin{tabular}{ll ccccc>{\columncolor{gray!15}}c ccccc>{\columncolor{gray!15}}c}
\toprule
\multirow{2}{*}{\textbf{Student}} & \multirow{2}{*}{\textbf{Method}} & \multicolumn{6}{c}{\textbf{Pass@1}} & \multicolumn{6}{c}{\textbf{Pass@8}} \\
\cmidrule(lr){3-8}\cmidrule(lr){9-14}
& 
& \multicolumn{1}{c}{\textbf{MATH500}} 
& \multicolumn{1}{c}{\textbf{AIME24}} 
& \multicolumn{1}{c}{\textbf{AIME25}} 
& \multicolumn{1}{c}{\textbf{AMC23}} 
& \multicolumn{1}{c}{\textbf{GSM8K}} 
& \multicolumn{1}{>{\columncolor{gray!15}}c}{\textbf{Avg.}} 
& \multicolumn{1}{c}{\textbf{MATH500}} 
& \multicolumn{1}{c}{\textbf{AIME24}} 
& \multicolumn{1}{c}{\textbf{AIME25}} 
& \multicolumn{1}{c}{\textbf{AMC23}} 
& \multicolumn{1}{c}{\textbf{GSM8K}} 
& \multicolumn{1}{>{\columncolor{gray!15}}c}{\textbf{Avg.}} \\
\midrule
\multicolumn{14}{c}{\textbf{\textit{Qwen3-Base Series (Teacher Size: 8B)}}} \\
\rowcolor{cordrow} 0.6B & Untrained & $12.7$ & $0.4$ & $0.2$ & $3.1$ & $24.5$ & \cellcolor{gray!15} $8.2$ & $42.0$ & $3.3$ & $1.7$ & $18.6$ & $67.5$ & \cellcolor{gray!15} $26.6$ \\
  & OPD & $15.4$ & $0.2$ & $0.0$ & $5.5$ & $30.0$ & \cellcolor{gray!15} $10.2$ & $46.9$ & $1.7$ & $0.0$ & $28.1$ & $71.8$ & \cellcolor{gray!15} $29.7$ \\
  & \METHOD{} & {\boldmath$18.3$} & {\boldmath$0.4$} & {\boldmath$0.2$} & {\boldmath$6.3$} & {\boldmath$35.8$} & \cellcolor{gray!15} {\boldmath$12.2$} & {\boldmath$50.9$} & {\boldmath$3.3$} & {\boldmath$1.7$} & {\boldmath$32.4$} & {\boldmath$77.4$} & \cellcolor{gray!15} {\boldmath$33.1$} \\
\rowcolor{cordrow} 1.7B & Untrained & $24.1$ & $0.4$ & $0.2$ & $7.2$ & $43.7$ & \cellcolor{gray!15} $15.1$ & $57.1$ & $3.3$ & $1.7$ & $35.0$ & $85.0$ & \cellcolor{gray!15} $36.4$ \\
  & OPD & $23.5$ & $0.4$ & {\boldmath$0.4$} & $8.6$ & $43.6$ & \cellcolor{gray!15} $15.3$ & $57.5$ & $3.3$ & {\boldmath$3.3$} & $44.5$ & $86.6$ & \cellcolor{gray!15} $39.0$ \\
  & \METHOD{} & {\boldmath$26.6$} & {\boldmath$0.8$} & $0.0$ & {\boldmath$11.7$} & {\boldmath$47.4$} & \cellcolor{gray!15} {\boldmath$17.3$} & {\boldmath$62.5$} & {\boldmath$5.9$} & $0.0$ & {\boldmath$47.3$} & {\boldmath$87.5$} & \cellcolor{gray!15} {\boldmath$40.6$} \\
\rowcolor{cordrow} 4B & Untrained & $31.2$ & $0.4$ & $0.0$ & $13.6$ & $58.2$ & \cellcolor{gray!15} $20.7$ & $64.7$ & $3.3$ & $0.0$ & $49.7$ & $93.6$ & \cellcolor{gray!15} $42.3$ \\
  & OPD & $30.9$ & $0.6$ & {\boldmath$1.5$} & $13.4$ & $57.1$ & \cellcolor{gray!15} $20.7$ & $68.1$ & $3.0$ & {\boldmath$10.1$} & $45.2$ & $94.4$ & \cellcolor{gray!15} $44.2$ \\
  & \METHOD{} & {\boldmath$32.8$} & {\boldmath$1.7$} & $0.8$ & {\boldmath$15.8$} & {\boldmath$60.5$} & \cellcolor{gray!15} {\boldmath$22.3$} & {\boldmath$69.7$} & {\boldmath$10.6$} & $5.9$ & {\boldmath$56.6$} & {\boldmath$95.1$} & \cellcolor{gray!15} {\boldmath$47.6$} \\
\midrule
\multicolumn{14}{c}{\textbf{\textit{Qwen3.5-Base Series (Teacher Size: 9B)}}} \\
\rowcolor{cordrow} 0.8B & Untrained & $13.1$ & $0.0$ & $0.0$ & $4.7$ & $23.6$ & \cellcolor{gray!15} $8.3$ & $39.1$ & $0.0$ & $0.0$ & $24.9$ & $63.7$ & \cellcolor{gray!15} $25.5$ \\
  & OPD & $15.9$ & {\boldmath$0.2$} & {\boldmath$0.2$} & $4.7$ & $29.3$ & \cellcolor{gray!15} $10.1$ & $42.5$ & {\boldmath$1.7$} & {\boldmath$1.7$} & $26.1$ & $68.4$ & \cellcolor{gray!15} $28.1$ \\
  & SKD & $15.9$ & $0.0$ & $0.0$ & {\boldmath$6.6$} & $30.9$ & \cellcolor{gray!15} $10.7$ & $42.5$ & $0.0$ & $0.0$ & {\boldmath$33.6$} & $69.8$ & \cellcolor{gray!15} $29.2$ \\
  & \METHOD{} & {\boldmath$17.3$} & {\boldmath$0.2$} & {\boldmath$0.2$} & $5.9$ & {\boldmath$32.0$} & \cellcolor{gray!15} {\boldmath$11.1$} & {\boldmath$45.3$} & {\boldmath$1.7$} & {\boldmath$1.7$} & $29.9$ & {\boldmath$70.3$} & \cellcolor{gray!15} {\boldmath$29.7$} \\
\rowcolor{cordrow} 2B & Untrained & $21.3$ & $0.8$ & $0.2$ & $8.4$ & $38.9$ & \cellcolor{gray!15} $13.9$ & $54.5$ & $6.7$ & $1.7$ & $40.4$ & $80.5$ & \cellcolor{gray!15} $36.7$ \\
  & OPD & $22.6$ & {\boldmath$0.4$} & $0.0$ & $10.5$ & $46.4$ & \cellcolor{gray!15} $16.0$ & $53.2$ & {\boldmath$3.3$} & $0.0$ & $38.2$ & $83.7$ & \cellcolor{gray!15} $35.7$ \\
  & OPD \small{(\text{160 steps})} & $21.5$ & $0.2$ & {\boldmath$0.6$} & $10.5$ & $42.8$ & \cellcolor{gray!15} $15.1$ & $51.3$ & $1.7$ & {\boldmath$5.0$} & $39.6$ & $80.5$ & \cellcolor{gray!15} $35.6$ \\
  & SKD & $22.8$ & $0.2$ & $0.2$ & $10.5$ & $47.1$ & \cellcolor{gray!15} $16.2$ & $52.2$ & $1.7$ & $1.7$ & $41.1$ & $83.4$ & \cellcolor{gray!15} $36.0$ \\
  & \METHOD{} & {\boldmath$24.8$} & $0.0$ & $0.2$ & {\boldmath$11.7$} & {\boldmath$47.7$} & \cellcolor{gray!15} {\boldmath$16.9$} & {\boldmath$56.3$} & $0.0$ & $1.7$ & {\boldmath$44.5$} & {\boldmath$84.9$} & \cellcolor{gray!15} {\boldmath$37.5$} \\
\rowcolor{cordrow} 4B & Untrained & $29.1$ & $0.4$ & $0.2$ & $13.0$ & $56.1$ & \cellcolor{gray!15} $19.7$ & $63.9$ & $3.3$ & $1.7$ & $47.9$ & $93.0$ & \cellcolor{gray!15} $42.0$ \\
  & OPD & $29.9$ & $0.8$ & {\boldmath$0.4$} & {\boldmath$16.9$} & $60.9$ & \cellcolor{gray!15} $21.8$ & $61.8$ & $6.7$ & {\boldmath$3.3$} & {\boldmath$52.7$} & $93.2$ & \cellcolor{gray!15} $43.5$ \\
  & \METHOD{} & {\boldmath$33.1$} & {\boldmath$1.3$} & {\boldmath$0.4$} & $16.3$ & {\boldmath$63.4$} & \cellcolor{gray!15} {\boldmath$22.9$} & {\boldmath$66.7$} & {\boldmath$8.4$} & {\boldmath$3.3$} & $50.9$ & {\boldmath$94.2$} & \cellcolor{gray!15} {\boldmath$44.7$} \\
\bottomrule
\end{tabular}}
\caption{\textbf{Main results on public math benchmarks.} For all six students, CoRD has higher average pass@1 and pass@8 than OPD in this evaluation. Results use 16 generated samples per problem (percent). Pass@$k$ uses the unbiased estimator, and Avg. is the unweighted mean across the five benchmarks, computed from unrounded benchmark scores; it may differ slightly from the mean of the displayed values due to rounding. All distilled models train for 80 steps except ``OPD (160 steps)'' (Section~\ref{sec:cost}); \METHOD{} uses $\alpha=0.6$ and SKD the published top-25 rule. Bold indicates the highest value among distilled models for each student and column, including ties. Pass@16 results for the three Qwen3 students are reported in Appendix Table~\ref{tab:app-pass16}.}
\label{tab:public}
\end{table*}

%% file: sections/table_difficulty.tex
\begin{table}[t]
\centering
\footnotesize
\setlength{\tabcolsep}{4.5pt}
\begin{tabular}{lcccc}
\toprule
\multirow{2}{*}{\textbf{Student}} & \multicolumn{4}{c}{\textbf{Untrained student solved $\boldsymbol{k}$ of 16}} \\
\cmidrule(lr){2-5}
& \parbox{0.9cm}{\centering$\boldsymbol{k{=}0}$} & \parbox{0.9cm}{\centering\textbf{1--3}} & \parbox{0.9cm}{\centering\textbf{4--7}} & \parbox{0.9cm}{\centering\textbf{8--16}} \\
\midrule
Qwen3-0.6B & +5.3 & \textbf{+6.2} & +0.5 & 0.0 \\
Qwen3-1.7B & \textbf{+3.0} & +2.8 & +0.7 & 0.0 \\
Qwen3-4B & \textbf{+4.2} & \textbf{+4.2} & -\,0.3 & 0.0 \\
\addlinespace[2pt]
Qwen3.5-0.8B & +0.8 & \textbf{+1.5} & +1.2 & +1.3 \\
Qwen3.5-2B & -\,0.5 & \textbf{+2.9} & +0.7 & -\,0.2 \\
Qwen3.5-4B & \textbf{+5.9} & +2.3 & +0.3 & +0.1 \\
\bottomrule
\end{tabular}
\caption{\textbf{Pass@16 gains by problem difficulty.} The table reports pass@16 gains of \METHOD{} over OPD (percentage points) after 80 steps on 1{,}919 public-benchmark problems, grouped by how many of 16 samples the untrained student solved. Bold marks each student's largest gain; for all six students, these occur at $k \leq 3$. Gaps are computed before rounding from per-problem estimates.}
\label{tab:difficulty}
\end{table}

%% file: sections/analysis.tex
\section{Learning in the Blind Spot}
\label{sec:analysis}

We now ask where the gains come from. The analyses in this section distill the instruction-tuned Qwen3-1.7B and Qwen3-4B-Base, called 1.7B and 4B below, from Qwen3-8B-Base on MATH \cite{hendrycks2021math} for 40 steps, otherwise with the recipe of Section~\ref{sec:experiments}.

\paragraph{Measuring the Blind Spot.}
Before training we freeze two sets of states for each student: states reached by the untrained student's own rollouts, and states reached when the untrained student's rollouts are coupled to the teacher ($\alpha=1$). We measure the reverse KL to the teacher on both sets before and after training and report closure, $100\,(1-\overline{\mathrm{KL}}_{\text{after}}/\overline{\mathrm{KL}}_{\text{before}})$, the percentage of the initial mean divergence that training removes (negative if it grows). At the start, the divergence on teacher-reached states is 2.4 times that on student-reached states for the 1.7B student and 1.5 times for the 4B student; for the three Qwen3 base students of Table~\ref{tab:public} it is 1.9, 1.7 and 1.5 times.

\input{sections/table_controls}

\begin{figure*}[t]
\centering
\includegraphics[width=0.99\textwidth]{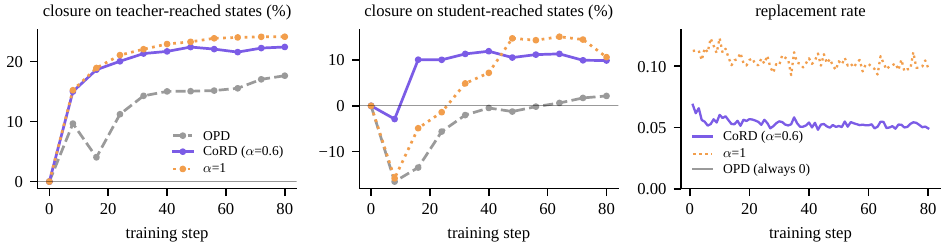}
\caption{\textbf{Teacher-side closure and replacement rate.} For the 4B-Base student and Qwen3-8B-Base teacher, \METHOD{} reaches in 16 steps the closure on teacher-reached states that OPD fails to reach in 80. Left and middle: closure of the reverse KL to the teacher on frozen teacher-reached and student-reached states, measured every 8 steps. Right: the realized replacement rate, which falls slowly as the student approaches the target.}
\label{fig:curves}
\end{figure*}

\paragraph{\METHOD{} Closes the Teacher-Side Gap.}
The first two rows of Table~\ref{tab:controls} show the effect of the sampler on the two sets. On teacher-reached states \METHOD{} closes 44.4\% of the gap at 1.7B against 37.4\% for OPD, and 21.7\% against 15.0\% at 4B. It does so without giving up the student's own states, where closure improves at 4B while OPD barely moves. Figure~\ref{fig:curves} follows the 4B student over 80 steps. The two curves on teacher-reached states separate within the first eight steps, and \METHOD{} reaches after 16 steps a closure that OPD does not reach in 80.

\paragraph{The Direction of the Replacement Matters.}
A natural objection is that any perturbation of the rollout would help by diversifying the training states. We test three alternatives. Two keep the acceptance test of \METHOD{} and replace the residual draw by a uniform random token or by a token drawn from the student's average token distribution. The third, in the spirit of DAgger, lets the teacher act at randomly chosen positions: at each position the teacher's own sample replaces the student's with a probability equal to \METHOD{}'s batch-mean replacement probability at that step, so the amount of intervention is comparable but its placement ignores where the two models disagree. At 1.7B all three intervene at least as often as \METHOD{} in Table~\ref{tab:controls}, yet none comes close: the random and frequency rules fall below OPD on both students, and the DAgger-style mixing stays within two points of it. Off-direction replacements steer the rollout into states that neither model prefers; at the first step the mean total-variation distance between student and teacher along the rollouts is 0.40 with uniform replacements against 0.15 with residual draws, so these rules spend the training signal on states that do not lead toward the teacher. The 4B student shows the same picture: at the first step a uniform replacement intervenes on 13.3\% of the tokens, twice \METHOD{}'s 6.9\%, and leaves the visited states twice as far from the teacher, with a mean total-variation distance of 0.22 against 0.11.

\input{sections/table_alpha}

\paragraph{Effect of $\boldsymbol{\alpha}$.}
\label{sec:alpha}
The parameter $\alpha$ sets how far the target leans toward the teacher, and with it how often the sampler intervenes. Table~\ref{tab:alpha} shows a clear trend: closure on teacher-reached states grows with $\alpha$ at both scales, and each tested $\alpha>0$ closes more of the teacher-side gap than OPD. On the public benchmarks after 80 steps, every $\alpha>0$ also improves pass@16 over OPD.

\paragraph{What the Student Gets Right.}
Accuracy can rise because the student formats answers more reliably or because the answers it commits to are more often correct. On the public benchmarks almost every sample ends in an extractable answer, 99.5 to 99.9\% for every run of Table~\ref{tab:public}, so the pass@1 gains of Table~\ref{tab:public} are gains in the correctness of the student's answers, not in their form. 

\paragraph{Interventions Recede With Training.}
Because the replacement probability equals the current total-variation distance, the sampler withdraws as the student approaches the target. Figure~\ref{fig:curves} shows that in the 80-step run at 4B, the replacement rate falls from 6.9\% at the first step to 4.9\% at step 80, without any schedule.

\paragraph{The Effect Holds Across Teachers.}
With Qwen3-14B as the teacher of the 1.7B student for 20 steps, \METHOD{} again closes more of the teacher-side gap than OPD, by 4.2 points, so the gain carries over to a second, larger teacher.

%% file: sections/table_controls.tex
\begin{table}[t]
\centering
\small
\setlength{\tabcolsep}{2pt}
\resizebox{0.99\columnwidth}{!}{%
\begin{tabular}{lrrrrr}
\toprule
\multirow{2}{*}{\textbf{Replacement Rule}} & \multirow{2}{*}{\parbox{1.3cm}{\centering\small{\textbf{Rate$_{\boldsymbol{1.7\text{B}}}$}}\\\centering\small{\textbf{(\%)}}}} & \multicolumn{2}{c}{\textbf{1.7B (Instruct)}} & \multicolumn{2}{c}{\textbf{4B-Base}} \\
\cmidrule(lr){3-4}\cmidrule(lr){5-6}
& & \textbf{Teach.} & \textbf{Stud.} & \textbf{Teach.} & \textbf{Stud.} \\
\midrule
none (OPD) & 0.0 & 37.4 & 52.0 & 15.0 & $-$0.5 \\
\rowcolor{cordrow} residual (\METHOD{}) & 7.5 & \textbf{44.4} & 52.0 & \textbf{21.7} & \textbf{11.9} \\
teacher sample, random position & 7.9 & 39.1 & 51.3 & 14.2 & $-$1.4 \\
average-distribution token & 12.2 & 29.2 & 48.4 & 13.7 & $-$10.4 \\
uniform token & 15.8 & 28.3 & 47.3 & 13.9 & $-$9.8 \\
\bottomrule
\end{tabular}}
\caption{\textbf{Gap closure under replacement rules.} Gap closure (\%) on frozen teacher-reached and student-reached states after 40 steps, for different replacement rules at $\alpha=0.6$. The random-position rule is a DAgger-style control whose average rate is close to \METHOD{}'s. Rate is the realized fraction of intervened positions at 1.7B. The residual draw is the only rule that clearly improves teacher-side closure for both students.}
\label{tab:controls}
\end{table}

%% file: sections/table_alpha.tex
\begin{table}[t]
\centering
\small
\setlength{\tabcolsep}{3.4pt}
\resizebox{0.99\columnwidth}{!}{%
\begin{tabular}{lrrrc}
\toprule
\multirow{2}{*}{$\boldsymbol\alpha$} & \multirow{2}{*}{\textbf{Rate (\%)}} & \multicolumn{2}{c}{\textbf{Teach. Closure}} & \textbf{4B, Public} \\
\cmidrule(lr){3-4}
& & \textbf{1.7B} & \textbf{4B} & \textbf{80 Steps} \\
\midrule
0 (OPD) & 0.0 & 37.4 & 15.0 & 20.7 / 50.3 \\
0.3 & 3.6 & 40.6 & 18.1 & 20.7 / 51.6 \\
\rowcolor{cordrow} 0.6 (\METHOD{}) & 7.5 & 44.4 & 21.7 & 22.3 / \textbf{54.9} \\
1 & 15.6 & \textbf{47.4} & \textbf{22.9} & \textbf{22.6} / 52.2 \\
\bottomrule
\end{tabular}}
\caption{\textbf{Closure and public accuracy across $\alpha$.} Closure on teacher-reached states grows with $\alpha$, and every $\alpha>0$ improves public pass@16 over OPD. Rate is the realized replacement rate at 1.7B; closure (\%) is measured on frozen teacher-reached states after 40 steps. Accuracy is pass@1 / pass@16 (\%) for the 4B-Base student, averaged over the five public benchmarks after 80 steps.}
\label{tab:alpha}
\end{table}

%% file: sections/conclusion.tex
\section{Conclusion}

On-policy distillation supervises the student only where the student already goes. \METHOD{} widens that set by sampling each rollout from the maximal coupling of the student and a teacher-leaning target: the student's token is kept where the two agree, and a token from the residual is substituted where they do not, exactly as often as they disagree. The loss, the data and the number of training steps stay as they are, and the method adds a single parameter: $\alpha=0$ is OPD exactly, and one value, $\alpha=0.6$, serves all six students. The interventions need no schedule, because the replacement rate falls by itself as the student approaches the target. Across six base students from two model families on public benchmarks, this change improves average pass@1 and pass@8 for all six students. Pass@16 gains are more variable, but each student's largest difficulty-bin gain occurs on problems it rarely or never solved before training. Frozen sets of teacher-reached states show that \METHOD{} closes more of the gap on exactly the states the teacher reaches. Because the sampler needs only the teacher's logits during generation, it combines with any change to the loss or its target.

%% file: sections/limitations.tex
\section*{Limitations}

\METHOD{} needs the teacher's logits during the rollout. Our public-benchmark experiments cover base students up to 4B parameters and teachers up to 9B (14B in one analysis), all on mathematical reasoning; larger models and other tasks are untested. Student and teacher must share a tokenizer. The AIME sets each contain only 30 problems, so small differences in pass rates should be interpreted cautiously.

\section*{Ethical Considerations}

This work studies model distillation using public mathematical data and benchmarks; it involves no human subjects or personal data. Improving a student's ability to follow a stronger model can also transfer that model's errors, biases, or unsafe behavior, and the resulting capabilities may be used outside the evaluated setting. Our experiments therefore should not be interpreted as a safety evaluation or as evidence that distilled models are reliable in high-stakes applications. We encourage evaluation for the intended deployment context, including independent accuracy and safety checks, before use.

%% file: appendix/details.tex
\clearpage
\section{Experimental Details}
\label{app:details}

\paragraph{Models.}
Public-benchmark experiments distill Qwen3-0.6B-Base, Qwen3-1.7B-Base and Qwen3-4B-Base from Qwen3-8B-Base \cite{yang2025qwen3}, and Qwen3.5-0.8B-Base, Qwen3.5-2B-Base and Qwen3.5-4B-Base from Qwen3.5-9B-Base \cite{qwen2026qwen35}. The analysis of Section~\ref{sec:analysis} distills Qwen3-1.7B (the instruction-tuned release) and Qwen3-4B-Base from Qwen3-8B-Base. All models are released under the Apache 2.0 license. Within each family the student and teacher tokenizers are checked to be identical before training.

\paragraph{Data.}
Public-benchmark runs train on the MATH training set \cite{hendrycks2021math} (MIT license): the first 5{,}000 problems of one fixed shuffle, minus the few (under 1\%) whose prompt exceeds 512 tokens, visited in the same order by every run (80 steps of 256 prompts cover the set about four times). The analysis runs train on the MATH training set \cite{hendrycks2021math} (MIT license) under one fixed shuffle: rows $[0, 5000)$ are used for training, rows $[5000, 5200)$ form the held-out set (199 problems after dropping prompts over 512 tokens), and rows from 6000 on seed the frozen state sets. Evaluation uses MATH500 \cite{lightman2024verify} (MIT), GSM8K \cite{cobbe2021gsm8k} (MIT, 1{,}319 test problems), and the AIME 2024, AIME 2025 and AMC 2023 problems \cite{aime2024, aime2025, amc2023} (30, 30 and 40 problems), which are publicly released competition problems used here for research evaluation only. A check of the 5{,}000 training problems against all 1{,}919 evaluation problems (normalized exact match, and 13-gram or character 5-gram near-duplicate screening followed by manual inspection) found one duplicate, a single MATH500 problem, and no overlap with AIME, AMC or GSM8K.

\paragraph{Training.}
Table~\ref{tab:app-hparams} lists the training and evaluation settings. Every run minimizes the full-vocabulary reverse KL $D_{\mathrm{KL}}(p_{\mathrm{S}} \,\|\, p_{\mathrm{T}})$ at every visited state, averaged over all tokens in the step, and draws rollout tokens from the full distribution, for OPD's student draws and for the coupled sampler alike. In the public-benchmark runs the frozen teacher is sharded across devices, which we checked to give logits identical to an unsharded teacher. Runs within a comparison differ only in the sampling rule; the data order and sampling randomness are held fixed, and the frozen state sets are shared by all runs of a student. Memory-saving options, where needed, are applied identically to every run of a student and were checked not to change any computed value. In the controls of Section~\ref{sec:analysis}, the average-distribution token is drawn from an exponential moving average (decay 0.9) of the student's next-token distribution over earlier rollout batches (the current batch's mean on the first batch).

\input{sections/table_app_hparams}

\paragraph{Prompt.}
Training and evaluation use the same two-shot completion prompt (Figure~\ref{fig:app-prompt}), with generation stopped at the next \texttt{Problem:}.

\begin{figure*}[htpb]
\centering
\begin{tcolorbox}[colback=black!5!white, colframe=black!75!black, title=Prompt for training and evaluation (two-shot completion)]
\small\ttfamily
Problem: What is the value of \$2\^{}3 + 3\^{}2\$?\\
Solution: We compute \$2\^{}3 = 8\$ and \$3\^{}2 = 9\$, so the sum is \$8 + 9 = 17\$.\\
The final answer is \$\textbackslash boxed\{17\}\$.\\[3pt]
Problem: If \$3x - 7 = 14\$, what is \$x\$?\\
Solution: Adding \$7\$ to both sides gives \$3x = 21\$, so \$x = 7\$.\\
The final answer is \$\textbackslash boxed\{7\}\$.\\[3pt]
Problem: \{question\}\\
Solution:
\end{tcolorbox}
\caption{The two-shot completion prompt used for training and for every evaluation; \texttt{\{question\}} is replaced by the problem, and generation stops at the next \texttt{Problem:}.}
\label{fig:app-prompt}
\end{figure*}

\input{sections/table_app_pass16}

\paragraph{Evaluation.}
On the public benchmarks the final \verb|\boxed{}| answer is compared with the reference by the SimpleRL-Zoo math equivalence checker \cite{zeng2025simplerlzoo} with a 3-second time limit (a time-out counts as incorrect). The held-out evaluations of Figure~\ref{fig:curves-public} score answers by normalized exact match between the reference and the final answer, read from the last \verb|\boxed{}| or, failing that, after ``The final answer is''. pass@$k$ is the unbiased estimator of \citet{chen2021codex} computed from all samples; maj@8 is the accuracy of the most frequent answer among 8 samples.

\paragraph{Compute.}
All experiments ran on accelerator hardware. For reference, the training cost was about 670 accelerator-hours for the twelve OPD and \METHOD{} configurations represented in Table~\ref{tab:public} (about 2 hours on 16 devices per Qwen3 run, and 5 to 19 hours on 8 devices per Qwen3.5 run), about 230 accelerator-hours for the training runs of Section~\ref{sec:analysis}, and about 170 accelerator-hours for the SKD runs and the 160-step OPD run of Section~\ref{sec:experiments}. Evaluation, which generates 16 samples for each of the 1{,}919 benchmark problems per model, is not included.

\section{Exactness of the Sampler}
\label{app:exact}
Before any training we checked the implementation against the closed form. On small vocabularies, the histogram of emitted tokens agrees with $q_\alpha$ for $\alpha \in \{0, 0.3, 0.6, 1\}$, and the realized rejection rate agrees with the analytic total-variation distance to within $3 \times 10^{-4}$; a sampler that replaces rejected tokens uniformly fails the same check by a wide margin. With the real 4B student and 8B teacher (200 prompts, 256 new tokens), the rejection rate agrees with the mean total-variation distance at every $\alpha$, the $\alpha=0$ sampler never rejects, and its gradient equals that of plain OPD exactly. During training of the 4B student ($\alpha=0.6$, 40 steps), the realized replacement rate stays within 0.003 of the mean total-variation distance computed from the same logits at every step. The SKD sampler of Section~\ref{sec:skd} was checked against its own closed form in the same way.

\section{Additional Results}
\label{app:extra}

\paragraph{Pass@16 on the Public Benchmarks.}
Table~\ref{tab:app-pass16} gives pass@16 for the three Qwen3 students of Table~\ref{tab:public}. \METHOD{} improves average pass@16 over OPD by 3.0 and 4.6 points for Qwen3-0.6B and Qwen3-4B, respectively; the Qwen3-1.7B result is 0.1 points lower.

%% file: sections/table_app_hparams.tex
\begin{table*}[t]
\centering
\small
\begin{tabular}{@{}l>{\centering\arraybackslash}p{4.6cm}>{\centering\arraybackslash}p{2.9cm}>{\centering\arraybackslash}p{2.9cm}@{}}
\toprule
 & \multicolumn{3}{c}{\textbf{Run group}} \\
\cmidrule(lr){2-4}
\textbf{Setting} & \textbf{Public benchmarks} & \textbf{Analysis on MATH} & \textbf{Qwen3-14B teacher} \\
 & (Table~\ref{tab:public}) & (Section~\ref{sec:analysis}) & (Section~\ref{sec:analysis}) \\
\midrule
\multicolumn{4}{l}{\textit{Training}} \\
Training steps & 80 (160 for ``OPD, 160 steps'') & 40 & 20 \\
Prompts per step & 256 & 256 & 64 \\
Max.\ new tokens per rollout & 320 & 320 & 224 \\
Rollout sampling & \multicolumn{3}{c}{temperature 1.0, top-$p$ 1.0} \\
Optimizer & \multicolumn{3}{c}{AdamW, $\beta = (0.9, 0.95)$, $\epsilon = 10^{-8}$, weight decay 0} \\
Learning rate & \multicolumn{3}{c}{$10^{-5}$, linear warm-up over 4 steps} \\
Gradient clipping & \multicolumn{3}{c}{global norm 1.0} \\
Precision & \multicolumn{3}{c}{bf16 mixed, fp32 optimizer states; frozen teacher in bf16} \\
\midrule
\multicolumn{4}{l}{\textit{Evaluation}} \\
Benchmark evaluation & 16 samples per problem, temperature 1.0, top-$p$ 1.0, at most 4{,}096 new tokens & -- & -- \\
In-run held-out (Figure~\ref{fig:curves-public}) & 8 samples per problem at steps 0 and 40, 16 at step 80, at most 384 new tokens & -- & -- \\
\bottomrule
\end{tabular}
\caption{Training and evaluation settings per run group. The analysis and Qwen3-14B-teacher runs are measured by closure on frozen state sets (Section~\ref{sec:analysis}).}
\label{tab:app-hparams}
\end{table*}

%% file: sections/table_app_pass16.tex
\begin{table*}[t]
\centering
\small
\setlength{\tabcolsep}{4pt}
\resizebox{0.8\textwidth}{!}{%
\begin{tabular}{llcccccc}
\toprule
\textbf{Student} & \textbf{Method} & \textbf{MATH500} & \textbf{AIME24} & \textbf{AIME25} & \textbf{AMC23} & \textbf{GSM8K} & \textbf{Avg.} \\
\midrule
\rowcolor{cordrow} Qwen3-0.6B-Base & Untrained & 53.8 & 6.7 & 3.3 & 30.0 & 79.5 & 34.7 \\
  & OPD & 57.6 & 3.3 & 0.0 & 45.0 & 81.6 & 37.5 \\
  & \METHOD{} & 62.6 & 6.7 & 3.3 & 45.0 & 85.1 & 40.5 \\
\addlinespace[2pt]
\rowcolor{cordrow} Qwen3-1.7B-Base & Untrained & 66.4 & 6.7 & 3.3 & 50.0 & 91.2 & 43.5 \\
  & OPD & 66.6 & 6.7 & 6.7 & 65.0 & 92.4 & 47.5 \\
  & \METHOD{} & 71.8 & 10.0 & 0.0 & 62.5 & 92.6 & 47.4 \\
\addlinespace[2pt]
\rowcolor{cordrow} Qwen3-4B-Base & Untrained & 72.6 & 6.7 & 0.0 & 62.5 & 96.3 & 47.6 \\
  & OPD & 77.0 & 3.3 & 16.7 & 57.5 & 97.0 & 50.3 \\
  & \METHOD{} & 77.6 & 16.7 & 10.0 & 72.5 & 97.8 & 54.9 \\
\bottomrule
\end{tabular}}
\caption{\textbf{Pass@16 (\%) on the five public benchmarks for the three Qwen3 base students.} Each entry is computed from 16 samples per problem, with Avg. the unweighted mean across benchmarks, computed from unrounded benchmark scores; it may differ slightly from the mean of the displayed values due to rounding. The Qwen3-1.7B average and some benchmark scores are lower with \METHOD{} than with OPD, despite higher average pass@1 and pass@8.}
\label{tab:app-pass16}
\end{table*}

%% file: appendix/sampler.tex
\section{Properties of the Coupled Sampler}
\label{app:sampler}

This appendix collects the properties of the target $q_\alpha$ and of the coupled sampler of Section~\ref{sec:method}, with complete proofs. Throughout, $p_{\mathrm{S}}$ and $p_{\mathrm{T}}$ are the student and teacher next-token distributions at one fixed state $s$, which we drop from the notation. Both are softmax outputs and therefore strictly positive on $\mathcal{V}$, and $\alpha \in [0,1]$. Table~\ref{tab:app-notation} lists the notation.

\begin{table}[h]
\centering
\small
\begin{tabular}{@{}ll@{}}
\toprule
Symbol & Meaning \\
\midrule
$p_{\mathrm{S}}, p_{\mathrm{T}}$ & student, teacher next-token distributions \\
$q_\alpha$ & tempered target of Eq.~\eqref{eq:target} \\
$Z_\alpha$ & normalizer $\sum_{u} p_{\mathrm{S}}(u)^{1-\alpha} p_{\mathrm{T}}(u)^{\alpha}$ \\
$\ell(v)$ & log ratio $\log p_{\mathrm{T}}(v) - \log p_{\mathrm{S}}(v)$ \\
$(x)_+$ & $\max(x, 0)$ \\
$\TV(p, q)$ & $\tfrac12 \sum_v |p(v) - q(v)|$ \\
\bottomrule
\end{tabular}
\caption{Notation for Appendix~\ref{app:sampler}.}
\label{tab:app-notation}
\end{table}

\paragraph{Exactness and Minimal Replacement.}
We first give the full proof of Proposition~\ref{prop:exact}. Let $a \sim p_{\mathrm{S}}$ be the proposal and $y$ the emitted token. The proposal is kept with probability $\min(1, q_\alpha(a)/p_{\mathrm{S}}(a))$, so the probability that $v$ is proposed and kept is $p_{\mathrm{S}}(v)\min(1, q_\alpha(v)/p_{\mathrm{S}}(v)) = \min(p_{\mathrm{S}}(v), q_\alpha(v))$. The total rejection probability is therefore
\begin{align*}
&\textstyle\sum_v \big(p_{\mathrm{S}}(v) - \min(p_{\mathrm{S}}(v), q_\alpha(v))\big) \\
&\quad= \textstyle\sum_v \big(p_{\mathrm{S}}(v) - q_\alpha(v)\big)_+ = \TV(p_{\mathrm{S}}, q_\alpha),
\end{align*}
where the last step uses $\sum_v p_{\mathrm{S}}(v) = \sum_v q_\alpha(v) = 1$. After a rejection the replacement is drawn from $(q_\alpha - p_{\mathrm{S}})_+ / \TV(p_{\mathrm{S}}, q_\alpha)$, so $v$ is emitted as a replacement with probability $(q_\alpha(v) - p_{\mathrm{S}}(v))_+$. Adding the two cases,
\begin{align*}
\Pr(y = v) &= \min(p_{\mathrm{S}}(v), q_\alpha(v)) \\
&\quad + (q_\alpha(v) - p_{\mathrm{S}}(v))_+ = q_\alpha(v). \quad\square
\end{align*}

\begin{lemma}[Minimal replacement]
\label{lem:maximal-coupling}
For every joint distribution of a pair $(a, y)$ with marginals $p_{\mathrm{S}}$ and $q_\alpha$, $\Pr(a \neq y) \geq \TV(p_{\mathrm{S}}, q_\alpha)$. The coupled sampler attains equality.
\end{lemma}

\begin{proof}
For every token $v$, $\Pr(a = y = v) \leq \min(p_{\mathrm{S}}(v), q_\alpha(v))$, so $\Pr(a = y) \leq \sum_v \min(p_{\mathrm{S}}(v), q_\alpha(v)) = 1 - \TV(p_{\mathrm{S}}, q_\alpha)$ \citep{lindvall1992coupling}. In the coupled sampler a proposal $a$ can be rejected only when $q_\alpha(a) < p_{\mathrm{S}}(a)$, where the residual $(q_\alpha - p_{\mathrm{S}})_+$ vanishes, so a replacement never equals the rejected proposal. Hence $\Pr(a \neq y)$ equals the rejection probability $\TV(p_{\mathrm{S}}, q_\alpha)$.
\end{proof}

\noindent Summed over the states of a rollout, the expected number of replacements is $\E\big[\sum_t \TV\big(p_{\mathrm{S}}(\cdot \mid s_t), q_\alpha(\cdot \mid s_t)\big)\big]$ over the visited states $s_t$, which are distributed as $d_{q_\alpha}$. Its per-token average, the mean total-variation distance, is the quantity that the realized replacement rate tracks in Appendix~\ref{app:exact}.

\paragraph{The Target as a Weighted Divergence Minimizer.}
The geometric interpolation of Eq.~\eqref{eq:target} has a variational description that also bounds its normalizer.

\begin{proposition}
\label{prop:variational}
For $\alpha \in [0,1]$, $q_\alpha$ is the unique minimizer over distributions $q$ on $\mathcal{V}$ of
\[
(1-\alpha)\, D_{\mathrm{KL}}(q \,\|\, p_{\mathrm{S}}) + \alpha\, D_{\mathrm{KL}}(q \,\|\, p_{\mathrm{T}}),
\]
and the minimum value is $-\log Z_\alpha$. Consequently $Z_\alpha \leq 1$, with equality for $\alpha \in (0,1)$ if and only if $p_{\mathrm{S}} = p_{\mathrm{T}}$.
\end{proposition}

\begin{proof}
Combining the two logarithms in each divergence,
\begin{align*}
&(1-\alpha)\, D_{\mathrm{KL}}(q \,\|\, p_{\mathrm{S}}) + \alpha\, D_{\mathrm{KL}}(q \,\|\, p_{\mathrm{T}}) \\
&\quad= \textstyle\sum_v q(v)\log\dfrac{q(v)}{p_{\mathrm{S}}(v)^{1-\alpha}p_{\mathrm{T}}(v)^{\alpha}} \\
&\quad= D_{\mathrm{KL}}(q \,\|\, q_\alpha) - \log Z_\alpha .
\end{align*}
The divergence is nonnegative and vanishes only at $q = q_\alpha$, which proves the first two claims. Both divergences in the objective are nonnegative, so $-\log Z_\alpha \geq 0$. For $\alpha \in (0,1)$ the minimum is zero only if a single $q$ makes both divergences vanish, that is, only if $q = p_{\mathrm{S}} = p_{\mathrm{T}}$.
\end{proof}

\begin{lemma}[Where the residual lives]
\label{lem:residual}
For $\alpha \in (0,1]$, $q_\alpha(v) > p_{\mathrm{S}}(v)$ if and only if $\ell(v) > \alpha^{-1}\log Z_\alpha$. In particular, if $\alpha \in (0,1)$ and $p_{\mathrm{S}} \neq p_{\mathrm{T}}$, every token with $p_{\mathrm{T}}(v) \geq p_{\mathrm{S}}(v)$ carries positive residual mass.
\end{lemma}

\begin{proof}
By Eq.~\eqref{eq:target}, $q_\alpha(v)/p_{\mathrm{S}}(v) = e^{\alpha\ell(v)}/Z_\alpha$, which exceeds one exactly when $\alpha\ell(v) > \log Z_\alpha$. Under the second hypothesis Proposition~\ref{prop:variational} gives $\log Z_\alpha < 0$, so $\ell(v) \geq 0$ suffices.
\end{proof}

\paragraph{Departure from the Student Grows with $\alpha$.}
The coefficient $\alpha$ controls how far the rollout distribution moves from the student at each state, and it does so monotonically.

\begin{proposition}
\label{prop:monotone}
Let $D(\alpha) = D_{\mathrm{KL}}(q_\alpha \,\|\, p_{\mathrm{S}})$. Then $D$ is nondecreasing on $[0,1]$, with $D(0) = 0$ and $D(1) = D_{\mathrm{KL}}(p_{\mathrm{T}} \,\|\, p_{\mathrm{S}})$, and the replacement probability satisfies
\[
\TV(p_{\mathrm{S}}, q_\alpha) \leq \sqrt{D(\alpha)/2} \leq \sqrt{D_{\mathrm{KL}}(p_{\mathrm{T}} \,\|\, p_{\mathrm{S}})/2}.
\]
\end{proposition}

\begin{proof}
Write $A(\alpha) = \log Z_\alpha = \log \sum_v p_{\mathrm{S}}(v) e^{\alpha\ell(v)}$, the cumulant generating function of $\ell$ under $p_{\mathrm{S}}$. Differentiating, $A'(\alpha) = \E_{q_\alpha}[\ell]$ and $A''(\alpha) = \mathrm{Var}_{q_\alpha}(\ell) \geq 0$. Since $\log(q_\alpha(v)/p_{\mathrm{S}}(v)) = \alpha\ell(v) - A(\alpha)$,
\[
D(\alpha) = \alpha A'(\alpha) - A(\alpha),
\]
so that $D'(\alpha) = \alpha A''(\alpha) \geq 0$.
The endpoint values follow from $q_0 = p_{\mathrm{S}}$ and $q_1 = p_{\mathrm{T}}$. The first inequality is Pinsker's inequality, and the second follows from monotonicity.
\end{proof}

\noindent Raising $\alpha$ therefore moves the target monotonically away from the student, and at no $\alpha$ does the sampler intervene with probability above $\sqrt{D_{\mathrm{KL}}(p_{\mathrm{T}} \,\|\, p_{\mathrm{S}})/2}$, a quantity fixed by the student--teacher gap at the state that vanishes as the student reaches the teacher.

\paragraph{The Worked Example in Full.}
Figure~\ref{fig:example} uses $p_{\mathrm{S}} = (0.6, 0.3, 0.1)$ and $p_{\mathrm{T}} = (0.2, 0.3, 0.5)$ on the tokens $(a, b, c)$ with $\alpha = 0.6$. Table~\ref{tab:app-example} lists every intermediate quantity.

\begin{table}[h]
\centering
\small
\setlength{\tabcolsep}{3.5pt}
\begin{tabular}{@{}lcccccc@{}}
\toprule
Token & $p_{\mathrm{S}}$ & $p_{\mathrm{T}}$ & $p_{\mathrm{S}}^{0.4}p_{\mathrm{T}}^{0.6}$ & $q_\alpha$ & Keep & Residual \\
\midrule
$a$ & 0.6 & 0.2 & 0.310 & 0.356 & 0.59 & 0 \\
$b$ & 0.3 & 0.3 & 0.300 & 0.344 & 1 & 0.044 \\
$c$ & 0.1 & 0.5 & 0.263 & 0.301 & 1 & 0.201 \\
\bottomrule
\end{tabular}
\caption{Every quantity of the worked example of Figure~\ref{fig:example} ($\alpha = 0.6$, $Z_\alpha = 0.873$). Keep is $\min(1, q_\alpha/p_{\mathrm{S}})$; Residual is $(q_\alpha - p_{\mathrm{S}})_+$.}
\label{tab:app-example}
\end{table}

The normalizer is $Z_\alpha = 0.873$, so the residual threshold of Lemma~\ref{lem:residual} is $Z_\alpha^{1/\alpha} = 0.80$ on the ratio $p_{\mathrm{T}}/p_{\mathrm{S}}$. Token $c$ (ratio 5) and token $b$ (ratio 1) lie above it and token $a$ (ratio $1/3$) lies below, so the residual has mass $0.044$ on $b$ and $0.201$ on $c$, which normalize to 18\% and 82\%. The replacement probability is $\TV(p_{\mathrm{S}}, q_\alpha) = 0.244$, all of it from proposals of $a$ ($0.6 \times 0.407$). Proposition~\ref{prop:monotone} is visible here as well: $D(0.6) = 0.19$ and $D_{\mathrm{KL}}(p_{\mathrm{T}} \,\|\, p_{\mathrm{S}}) = 0.58$, so the replacement probability $0.244$ lies below both bounds, $0.31$ and $0.54$. Across $\alpha \in \{0, 0.3, 0.6, 1\}$ the replacement probability is $0$, $0.12$, $0.24$ and $0.40$, and $D(\alpha)$ is $0$, $0.04$, $0.19$ and $0.58$. By contrast, a rule that replaces every token outside the teacher's top-1 with a teacher sample intervenes whenever $a$ or $b$ is proposed, with probability $0.9$, and its replacement is $a$ with probability $p_{\mathrm{T}}(a) = 0.2$.

\paragraph{Computation in Log Space.}
Every quantity of Algorithm~\ref{alg:cord} follows from the two log-probability vectors at the current position. The target is $\log q_\alpha = (1-\alpha)\log p_{\mathrm{S}} + \alpha\log p_{\mathrm{T}} - \log Z_\alpha$, with $\log Z_\alpha$ a log-sum-exp over $\mathcal{V}$; the keep test is $\log u \leq \log q_\alpha(a) - \log p_{\mathrm{S}}(a)$; and the residual weights are $(q_\alpha - p_{\mathrm{S}})_+$, whose sum is the rejection probability. The sampler therefore adds no forward pass beyond the two that the loss of Eq.~\eqref{eq:opd} already requires.

\paragraph{The Two Endpoints.}
At $\alpha = 0$ the target is the student, $Z_0 = 1$, and no proposal is rejected, so the rollout is OPD's. At $\alpha = 1$ the target is the teacher and the sampler is exactly the speculative sampler of \citet{leviathan2023speculative} with the student as the draft model: every emitted token is a teacher sample, and the replacement probability is $\TV(p_{\mathrm{S}}, p_{\mathrm{T}})$. Intermediate values of $\alpha$ interpolate between these two rollout distributions along the path of Proposition~\ref{prop:variational}.